\pdfoutput=1

\documentclass[11pt]{article}

\usepackage[]{EMNLP2023}

\usepackage{times}
\usepackage{latexsym}

\usepackage[T1]{fontenc}

\usepackage[utf8]{inputenc}

\usepackage{microtype}

\usepackage{inconsolata}

\usepackage[T2A,T1]{fontenc}

\usepackage[utf8]{inputenc}

\usepackage{microtype}

\newcommand{\nc}[1]{\textcolor{orange}{Nicola: #1}}

\usepackage{url}

\usepackage{amsmath,amsfonts,bm}

\def\eqref#1{equation~\ref{#1}}

\def\1{\bm{1}}

\DeclareMathAlphabet{\mathsfit}{\encodingdefault}{\sfdefault}{m}{sl}
\SetMathAlphabet{\mathsfit}{bold}{\encodingdefault}{\sfdefault}{bx}{n}

\DeclareMathOperator*{\argmax}{arg\,max}

\usepackage{hyperref}
\usepackage{graphicx}
\usepackage{caption}
\usepackage{subcaption}
\usepackage{physics}
\usepackage{enumitem}
\usepackage{booktabs}
\usepackage{mathtools}
\usepackage{footmisc}
\usepackage{makecell}
\usepackage{xcolor}
\usepackage{xspace}
\newcounter{srCounter}
\newif\ifsrvar
\srvartrue
\ifsrvar
\newcommand{\seb}[1]{{\small \color{red} \refstepcounter{srCounter}\textsf{[SR]$_{\arabic{srCounter}}$:{#1}}}}
\else
\newcommand{\seb}[1]{}
\fi

\newcounter{fpCounter}
\newif\iffpvar
\fpvartrue
\iffpvar
\newcommand{\fabio}[1]{{\small \color{blue} \refstepcounter{fpCounter}\textsf{[FP]$_{\arabic{fpCounter}}$:{#1}}}}
\else
\newcommand{\fabio}[1]{}
\fi

\newcounter{trCounter}
\newif\iftrvar
\trvartrue
\iftrvar
\newcommand{\tim}[1]{{\small \color{purple} \refstepcounter{trCounter}\textsf{[TR]$_{\arabic{trCounter}}$:{#1}}}}
\else
\newcommand{\tim}[1]{}
\fi

\newcounter{apCounter}
\newif\ifapvar
\apvartrue
\ifapvar
\newcommand{\piktus}[1]{{\small \color{orange} \refstepcounter{apCounter}\textsf{[AP]$_{\arabic{apCounter}}$:{#1}}}}
\else
\newcommand{\piktus}[1]{}
\fi

\newcounter{plCounter}
\newif\ifplvar
\plvartrue
\ifplvar
\newcommand{\patrick}[1]{{\small \color{magenta} \refstepcounter{plCounter}\textsf{[PL]$_{\arabic{plCounter}}$:{#1}}}}
\else
\newcommand{\patrick}[1]{}
\fi

\newcounter{afCounter}
\newif\ifafvar
\afvartrue
\ifafvar
\newcommand{\angela}[1]{{\small \color{olive} \refstepcounter{afCounter}\textsf{[AF]$_{\arabic{afCounter}}$:{#1}}}}
\else
\newcommand{\angela}[1]{}
\fi

\newcounter{kpCounter}
\newif\ifkpvar
\kpvartrue
\ifkpvar
\newcommand{\kpopat}[1]{{\small \color{red} \refstepcounter{kpCounter}\textsf{[KP]$_{\arabic{kpCounter}}$:{#1}}}}
\else
\newcommand{\kpopat}[1]{}
\fi

\newcounter{ndcCounter}
\newif\ifndcvar
\ndcvartrue
\ifndcvar

\else
\newcommand{\ndc}[1]{}
\fi

\newcounter{ncCounter}
\newif\ifncvar
\ncvartrue
\ifncvar

\else
\newcommand{\nc}[1]{}
\fi

\newcounter{lwCounter}
\newif\ifncvar
\ncvartrue
\ifncvar
\newcommand{\ledell}[1]{{\small \color{brown} \refstepcounter{lwCounter}\textsf{[LW]$_{\arabic{lwCounter}}$:{#1}}}}
\else
\newcommand{\ledell}[1]{}
\fi

\newif\ifncvar
\ncvartrue
\ifncvar
\newcommand{\mikep}[1]{{\small \color{green} \refstepcounter{ncCounter}\textsf{[MIKEP]$_{\arabic{ncCounter}}$:{#1}}}}
\else
\newcommand{\mikep}[1]{}
\fi

\newcounter{smCounter}
\newif\ifsmvar
\smvartrue
\ifsmvar

\else
\newcommand{\kpopat}[1]{}
\fi

\def\bela{\textsc{BELA}\@\xspace}

\title{Multilingual End to End Entity Linking}

\author{Mikhail Plekhanov, Nora Kassner, Kashyap Popat, \\ \textbf{Louis Martin, Simone Merello, Borislav Kozlovskii, Fr\'ed\'eric A. Dreyer, Nicola Cancedda} \\
Meta AI
\\
\texttt{movb@meta.com}
}

\begin{document}
\maketitle
\begin{abstract}
Entity Linking is one of the most common Natural Language Processing tasks in practical applications, but so far efficient end-to-end solutions with multilingual coverage have been lacking, leading to complex model stacks.
To fill this gap, we release and open source BELA, the first fully end-to-end multilingual entity linking model that efficiently detects and links entities in texts in any of 97 languages.
We provide here a detailed description of the model and report BELA's performance on four entity linking datasets covering high- and low-resource languages. %

\end{abstract}

\section{Introduction} \label{sec:intro}

\begin{figure*}
\centering
     \includegraphics[clip,width=0.9\textwidth]{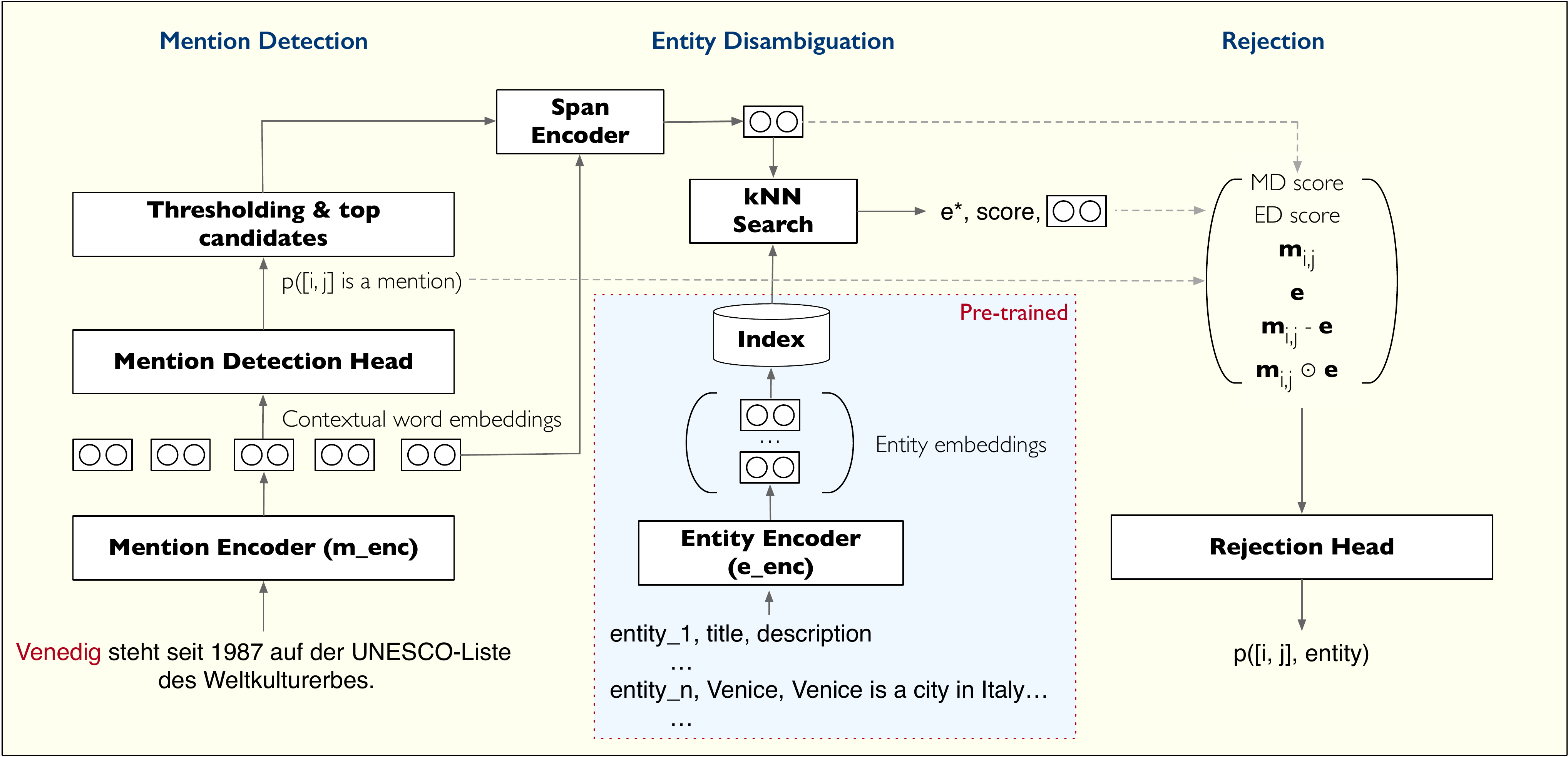}	   %
\caption{\textbf{Model architecture}: BELA is composed of a Mention Detection (MD), Entity Disambiguation (ED) and Rejection head (R). MD detects entity mentions in context by scoring potential start and end tokens and computes a mention encoding by pooling across mention tokens. Next, ED disambiguates the mention by running a kNN search between the mention encoding and encodings of candidate entities (the entity index). Entity encodings are computed on the basis of Wikipedia titles and abstracts. The top-ranked entity candidate is passed to R that selects only known and correctly linked entities and rejects everything else. See section~\ref{sec:model} for detailed description.
\label{architecture}}

\end{figure*}

Entity Linking (EL) consists in connecting referring expressions in text to the corresponding real-world entities, and is therefore a fundamental task in Natural Language Processing. It is an essential component in Natural Language Understanding (NLU) stacks powering many downstream applications such as information extraction, content moderation and question answering \cite{6823700,yin-etal-2016-simple}.

Besides exhibiting high precision, a practical EL system must be fast and high-coverage.
This work addresses these requirements and introduces BELA  (Bi-encoder Entity Linking Architecture), the first transformer-based, end-to-end, one-pass, multilingual EL system, covering 97 languages.

BELA combines and extends progress that EL systems have made in recent years across several dimensions:

\begin{itemize}
    \item End-to-end but monolingual architectures that combine Mention Detection (MD) with Entity Disambiguation (ED) and can be trained jointly \cite{kolitsas-etal-2018-end}.
    \item Transformer-based architectures, which achieved significant performance improvements on the ED task in the monolingual \cite{wu-etal-2020-scalable} and in the highly multilingual \cite{botha-etal-2020-entity} problem setting.
    \item Single-pass end-to-end architectures built on top of transformers that have been proven successful on short text spans \cite{li2020elq}.
\end{itemize}

BELA meets efficiency demands by adopting a bi-encoder architecture that requires a single forward pass through a transformer for end-to-end linking of a passage, irrespective of the number of entity mentions appearing in it. For the core Entity Disambiguation sub-task, it runs a k-nearest neighbor (kNN) search using an encoded mention as query in an entity index, as in BLINK \cite{Wu2020ScalableZE}. %
Following \citet{botha-etal-2020-entity}, mentions in any language are linked to a language-agnostic entity index, enabling high coverage of approximately 16 million entities and 97 languages.

Overall, BELA has a measured throughput of 53 samples per second on a single GPU. Thanks to its efficiency, multilinguality and end-to-end nature, BELA can improve and also simplify entity-centric applications that would otherwise require complex routing logics and model combinations.

We release BELA's codebase and model in the hope that it will be useful in a variety of applications and will foster further research. \footnote{\url{https://github.com/facebookresearch/BELA}}

\section{Task}
We formally define multilingual, end-to-end EL as a natural extension of \citet{botha-etal-2020-entity}: given a sample paragraph $p$ in any language $l_{p}$ and an entity index consisting of a set of entities
$E = \{e_{i}\}$, each with canonical names, titles $t(e_{i})$,
and textual descriptions $d(e_{i})$, our goal is to output a list of tuples, $(e, [i, j])$, where $e \in E$ is the entity corresponding to the mention $m_{i,j}$ spanning from the $i^{th}$ to $j^{th}$ token in $p$. Note that an entity part of $E$ can be associated with a set of languages $L= \{l_{i}\}$, but $l_{p}$ does not have to overlap with $L$: BELA can recognise an entity in a language even if the only available descriptions are in other languages. Details on the multilingual construction of the entity index follow in the next section.

\section{Entity Index}
To construct the entity index, we follow prior work \cite{botha-etal-2020-entity} and rely on Wikipedia and WikiData \citep{vrandevcic2014wikidata}. WikiData is closely integrated with Wikipedia and covers a large set of entities spanning over multiple languages. In total the entity index consists of around 16M entities representing the union of all Wikipedia entities in 97 languages. We use Wikipedia titles as $t(e_{i})$ and the first paragraph of the article as $d(e_{i})$. We use the same data-driven selection heuristic as \citet{botha-etal-2020-entity} to select a single description language $l_{i} \in L$ for each entity based on the number of mentions of each entity in different languages.
\begin{figure*}
\centering
     \includegraphics[clip,width=0.8\textwidth]{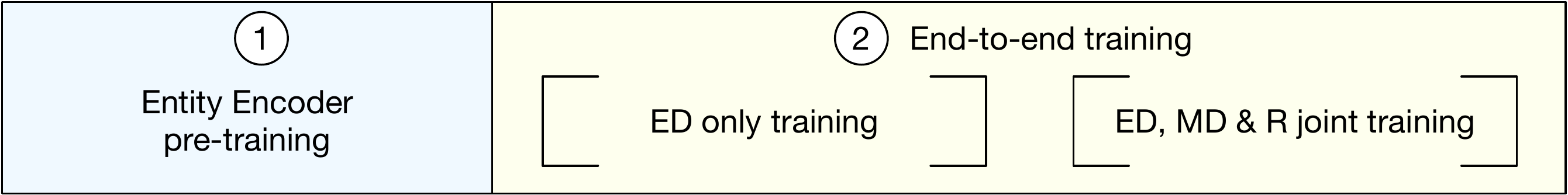}	   %
\caption{\textbf{Training steps}: Training BELA involves multiple steps: we first pre-train a bi-encoder ED-only model following BLINK (blue). We then freeze the pre-trained entity encoder and train the end-to-end model (yellow). See Figure \ref{architecture} for the the model parts corresponding to the different training steps.
\label{trainingsteps}}

\end{figure*}

\section{Model}
\label{sec:model}

BELA is a multilingual and end-to-end extension of the dense-retrieval-based models, e.g., BLINK \cite{Wu2020ScalableZE} and \citet{botha-etal-2020-entity}. It is composed of a Mention Detection (MD), Entity Disambiguation (ED) and Rejection (R) head. The MD and ED heads are built on top of encoders based on XLM-R-large \cite{lample2019cross} and fine-tuned to detect the boundaries $[i, j]$ of a mention in context and link it to an entity. ED follows BLINK's dense-retrieval approach that uses a bi-encoder architecture and runs a kNN search between the mention encoding and candidate entity encodings (the entity index). Finally, R decides whether to keep or discard each association between a mention and an entity.

An overview of the architecture is shown in Figure~\ref{architecture}. In the following section, we provide a detailed description of BELA's architecture. In Section~\ref{train}, we explain the training procedure in detail.

\subsection{Mention Detection}
Following \citet{li2020elq}, for every span $[i,j]$, the MD head calculates the probability of $[i,j]$ being the mention of an entity by scoring whether $i$ is the start position, $j$ is the end position, and the tokens between $i$ and $j$ are inside the mention:%
\begin{align*}
& s_{\mathrm{start},i} = \textbf{w}_{\mathrm{start}}^{\top}  \textbf{p}_{i}\,, \\
& s_{\mathrm{end},j} = \textbf{w}_{\mathrm{end}}^{\top}  \textbf{p}_{j}\,, \\
& s_{\mathrm{inside},t} = \textbf{w}_{\mathrm{inside}}^{\top}  \textbf{p}_{t}\,,
\end{align*}
where $\textbf{w}_\mathrm{start}, \textbf{w}_\mathrm{end}$, and $\textbf{w}_\mathrm{inside}$ are learnable vectors and $\textbf{p}_{i}$ token representations from the mention encoder:
\begin{align*}
[\textbf{p}_{1} \ldots \textbf{p}_{n}] = \mathrm{enc\_m}(p_{1} \ldots p_{n})\,,
\end{align*}
where $\mathrm{enc\_m}$ refers to a fine-tuned XLM-R encoder. We finally compute MD probability scores:
\begin{align*}
p([i, j]) = \sigma(s_{\mathrm{start},i} + s_{\mathrm{end},j} + \sum_{t=i+1}^{j-1} (s_{\mathrm{inside},t}))\,.
\end{align*}
We select candidates with $p>\beta$ as mention candidates and propagate them to the next step. $\beta$ is a mention detection hyperparameter that is set to 0.5 during training and then for testing tuned on a development set. For details see Appendix \ref{hyperparameters}.

\subsection{Entity Disambiguation}
The ED head receives an encoded candidate mention and finds the best match in the index. As in BLINK, our ED model is based on dense retrieval, i.e. the encoded mention is used as query in a kNN search among entity embeddings.

Entity embeddings are obtained by encoding canonical entity names and descriptions using the entity encoder:
\begin{align*}
\textbf{e} = \mathrm{enc\_e}(t(e), d(e))\,,
\end{align*}
where the entity encoder $\mathrm{enc\_e}$ is obtained by fine-tuning an XLM-R model as described in Sec.~\ref{train}.

We follow \citet{li2020elq} in constructing mention representations with one pass of the encoder and without mention boundary tokens by pooling mention tokens from the encoder output through a feed-forward layer (FFL):
\begin{align*}
\textbf{m}_{i,j} = FFL(\textbf{p}_{i} \ldots  \textbf{p}_{j})\,.
\end{align*}

We compute a similarity score $s$ between the mention candidate and an entity candidate $e \in E$ which we call the ED score:
\begin{align*}
s(e, [i, j]) = \textbf{e}^\top\textbf{m}_{i,j}\,.
\end{align*}

Finally we select $([i,j],e^*)$ where
\begin{align*}
e^* = \argmax _{e}(s(e,[i,j])),
\end{align*}
and pass it as a candidate $\langle\,\mathrm{mention\_span}, \mathrm{entity}\,\rangle$ tuple to a rejection head.

\subsection{Rejection head}

The MD and ED steps are intended to attain high recall, and therefore over-generate candidates. 
We introduce a rejection head R that looks at a $([i, j], e^*)$ pair holistically and decides whether to accept it, giving control over the precision-recall trade off. Input features to R are the MD score $p([i, j])$, the ED score $s(e^*, [i, j])$, the mention representation $\textbf{m}_{i,j}$, the top-ranked candidate representation $\textbf{e}^{*}$, as well as their difference and Hadamard product. The concatenation of these features is fed through a feed-forward network to output the final entity linking score $s([i, j], e^*)$. %
All $s([i, j], e^*)>\gamma$ are accepted where $\gamma$ is a threshold tuned on a development set to control between precision and recall. For details see Appendix \ref{hyperparameters}.

 \label{sec:results}
\begin{table*}[t!]
\centering
{
\begin{tabular}{lccccccccc} \hline
Model & ar & de & en & es & fa & ja & sr & ta & tr \\ \hline
 \textbf{F1} & 43.5 & 55.6 & 44.1 & 47.8 & 41.6 & 48.8 & 74.2 & 45.7 & 49.0 \\
 \textbf{R} & 42.6 & 50.0 & 40.3 & 42.3 & 35.9 & 43.8 & 68.7 & 41.5 & 49.5 \\
 \textbf{P} & 44.5 & 62.7 & 48.7 & 54.9 & 49.5 & 55.0 & 80.8 & 50.9 & 48.5 \\
 \bottomrule
\end{tabular}
}
\caption{End-to-end EL performance on Mewsli-9'-test (under-labeled for end-to-end linking). %
}
\label{Results-end2end}
\end{table*}

\begin{table*}[t!]
\centering
{
\begin{tabular}{lccccccccc} \hline
Model & ar & de & en & es & fa & ja & sr & ta & tr \\ \hline
  \textbf{F1} & 64.5 & 73.6 & 73.2 & 78.5 & 63.8 & 66.1 & 65.6 & 55.6 & 70.2 \\
 \textbf{R} & 84.4 & 85.5 & 82.8 & 85.6 & 78.0 & 74.5 & 87.8 & 67.5 & 77.2 \\
 \textbf{P} & 52.2 & 64.5 & 65.6 & 72.4 & 54.0 & 59.4 & 52.3 & 47.2 & 64.4 \\ 
 \bottomrule
\end{tabular}
}
\caption{End-to-end EL performance on Mewsli-9'-test (labeled for end-to-end linking).
}
\label{Results-end2end-silver}
\end{table*}

\section{Training}
\label{train}
Training BELA involves multiple steps: we first pre-train a bi-encoder ED-only model following BLINK. We then freeze the pre-trained entity encoder and train the end-to-end model. See Figure \ref{trainingsteps} for an overview of the training pipeline where the two training steps are color-coded to match their corresponding model components in Figure \ref{architecture}.

\subsection{Pre-trained entity encodings}
We pre-train the bi-encoder ED model on a Wikipedia dataset consisting of 661M samples. Wikipedia-based datasets are constructed from cross-article hyperlinks, mapping the anchor text to mentions and the link target to the corresponding entity. We consider all the Wikipedia articles spanning across 97 languages. A list of all languages can be found in the Appendix \ref{sec:lang}. 

We initialize the entity and mention encoder with XLM-R and train using softmax loss %
to maximize similarity between mention encoding and gold entity. %
We run two training iterations: a first one with in-batch random negatives and a second one using hard-negative training \cite{gillick-etal-2019-learning}. For training details on the entity encoder, we refer to \citet{Wu2020ScalableZE} and \citet{botha-etal-2020-entity}. 

\subsection{Training the end-to-end model}
Next, we freeze the pre-trained entity encoder and train the end-to-end model. Specifically, the mention encoder has to be adapted for end-to-end linking. We again proceed in two-steps. First, we run another round of ED-only training on Wikipedia data with 104M samples. Then, MD, ED, and R are trained jointly on another subset of Wikipedia data using 27M samples. Outputs from one component are fed as input to the next and losses are summed together. We use negative log-likelihood as our ED loss. A likelihood distribution over positive and mined hard negative entities \cite{gillick-etal-2019-learning} is computed for each mention. Using pre-trained entity encodings enables us to index entity embeddings using quantization algorithms for fast kNN search (using the FAISS \cite{JDH17} framework with the HNSW index). %

The MD loss is the binary cross-entropy between all possible valid spans and gold mentions in the training set. Valid spans are spans with $\mathrm{begin} < \mathrm{end}$, shorter than 256 tokens, and neither starting nor ending in the middle of a word.

The R head loss is the binary cross-entropy between retrieved and gold mention-entity pairs.

\section{Experimental Setting}

\subsection{Datasets}
\noindent

For evaluating \bela, we consider multiple benchmarking datasets spanning high- and low-resource languages:
\begin{itemize}
    \item \textbf{Mewsli-9}: An ED benchmarking data released by \citet{botha-etal-2020-entity}. This dataset contains manually labelled WikiNews articles in 9 different languages. We split this dataset in half to use half of the data for an additional round of fine-tuning. We call the data splits Mewsli-9'-train and -test. We will publish the corresponding data indices. 
    \item \textbf{TAC-KBP2015}: An ED dataset based on a set of topic-focused news articles and discussion forum posts released in the TAC-KBP2015 Tri-Lingual Entity Linking Track \citep{ji2015overview}. Following \cite{de-cao-etal-2022-multilingual}, we only consider Spanish and Chinese languages.
    \item \textbf{LORELEI}: An ED dataset based on a collection of manually labelled datasets focusing specifically on low resource languages \citep{strassel-tracey-2016-lorelei}.
    \item \textbf{AIDA}: A standard English end-to-end EL dataset based on YAGO2 KB \cite{hoffart-etal-2011-robust} used to benchmark English end-to-end performance.
\end{itemize}

Dataset statistics are summarized in Appendix Table~\ref{tab:data_stats}. Models evaluated on Mewsli-9'-test are finetuned on Mewsli-9'-train. Models evaluated on TAC-KBP2015, LORELEI and AIDA are finetuned on AIDA train.

\subsection{Evaluation Metrics}
For end-to-end performance, we report \textbf{F1} scores, recall (\textbf{R}) and precision (\textbf{P}) following \citet{li2020elq} but using a hard matching criteria. Whenever the input text is longer than the model maximum sequence length, we run the model on overlapping text windows of maximum sequence length, where the overlap size can be fine-tuned on a validation dataset. If several mentions overlap, we choose one with the highest MD score. For MD, we report recall where a mention is correctly detected if the start and end index match the gold label. For ED we report accuracy where a prediction is correct if the ground-truth entity is identified.

\begin{table*}[h]
\centering
{
\begin{tabular}{lccccccccc|c} 
\toprule
Model & ar & de & en & es & fa & ja & sr & ta & tr & avg. \\ 
\midrule
\citet{botha-etal-2020-entity} &  92.0 & 92.0 & 87.0 & 89.0 & 92.0 & 88.0 & 93.0 & 88.0 & 88.0 & 89.88 \\
\citet{de-cao-etal-2022-multilingual} &94.7 &91.5 &86.7 &90.0 &94.6 &89.9 &94.9 &92.9 &90.7 & 91.76
\\
\midrule
\textbf{BELA} & 90.0 & 90.0 & 85.0 & 88.0 & 91.0 & 85.0 & 93.0 & 90.0 & 87.0 & 88.77 \\
\bottomrule
\end{tabular}
}
\caption{ED accuracy on Mewsli-9.}
\label{Results-ED}
\end{table*}
\section{Results}
In the following sections we report results. We first present end-to-end performance. Next, we focus on the three components of the model individually, namely the MD, ED and R heads. We close with a discussion of computational complexity.

\subsection{End-to-end Results}

We report end-to-end performance for Mewsli-9'-test in Table \ref{Results-end2end}, for TAC-KBP in Table \ref{Results-tac} and for LORELEI in Table \ref{Results-lorelei}. Performance varies across languages between 15 to 74 F1 score. Such large variance could be due to a combination of the following reasons: i) Varying language coverage at training, e.g., \textit{sr} is most abundant in  Mewsli-9' which correlates with performance, ii) the tokenizer's performance being skewed towards some scripts and iii) general language dependent performance differences on the side of the underlying language model.

Note that as Mewsli-9', LORELEI and TAC-KBP are ED datasets, only a subset of the mentions are labelled. This results in over-estimating false positives, therefore under-estimating precision, when evaluating end-to-end EL. We illustrate this problem in Appendix \ref{Underlabelling}. A similar problem of under-labelling, here due to the incompleteness of Wikipedia hyperlinks, also impacts training: we hypothesize that our model underestimates the base rate at which it should detect mentions, impacting recall. 

We show that performance figures are higher if the model is trained on data exhaustively labelled with linked mentions. To this end, we create end-to-end benchmark data based on Wikipedia and Mewsli-9'. We use a model that was trained on a labelled non-public dataset to automatically label this data. We repeat training and evaluation and find that end-to-end performance increases significantly, see Table \ref{Results-end2end-silver}. We publish these labels as stand-off annotations for future reference.

As BELA is the first multilingual end-to-end EL system, we cannot compare to any baseline models in the multilingual setting.

To still provide some insight on its end-to-end performance in reference to baseline models, we report performance on the English end-to-end EL dataset AIDA in Table \ref{aida}. Note, that our entity index is computed on recent Wikipedia snapshots which misses 8\% of AIDA entities which were outdated. Overall we find that BELA performs competitively with English-only models, despite being highly multilingual and lacking sophisticated cross-attention between mentions and entity description, which would make the system impractically slow.

\begin{table}[t]
\centering
\begin{tabular}{lc}
\toprule
\textbf{Model} & \textbf{F1}  \\
\midrule
\citet{kolitsas-etal-2018-end}              & 82.4 \\
\citet{peters-etal-2019-knowledge}          & 73.7 \\
\citet{broscheit-2019-investigating}        & 79.3 \\
\citet{martins-etal-2019-joint}             & 81.9 \\
\citet{fevry2020empirical}                  & 76.7 \\
\citet{decao2021autoregressive}             & 83.7 \\
\midrule
\textbf{BELA} & 74.5\\
\bottomrule
\end{tabular}

\caption{End-to-end results on AIDA.}
\label{aida}
\end{table}

\begin{table*}[t!]
\centering
{
\begin{tabular}{lccccccccc} \hline
Model & ar & de & en & es & fa & ja & sr & ta & tr \\ \hline
 \textbf{R} & 87.5 & 91.4 & 89.1 & 91.6 & 86.1 & 81.4 & 92.6 & 72.5 & 87.0 \\
 \bottomrule
\end{tabular}
}
\caption{MD performance on Mewsli-9'-test (labeled for end-to-end linking).
}
\label{MD}
\end{table*}

\subsection{Mention Detection}

We report MD recall on Mewsli-9'-test in Table \ref{MD} using the model trained and tested with exhaustively labelled data. Recall varies between 73\% and 93\% across languages.

\subsection{Entity Disambiguation}

\begin{table}[t!]
\small
\centering
{
\begin{tabular}{lccc} \hline
Model & & TAC-KBP\\ \hline

& es & zh & macro-avg \\ \hline
\textbf{F1} & 64.8 &  52.4 & 58.6 \\
\textbf{R} & 64.1 & 49.0 & 56.6 \\
\textbf{P} & 65.6 & 56.4 & 61.0 \\
\bottomrule
\end{tabular}
}
\caption{End-to-end performance on TAC-KBP (under-labeled for end-to-end linking).
}
\label{Results-tac}
\end{table}

\begin{table}[t!]
\small
\centering
{
\begin{tabular}{lccccc} \hline
Model & & & LORELEI \\ \hline

 & vi & uk & ti & om & macro-avg \\ \hline
\textbf{F1} & 52.2 & 48.7 & 28.7 & 15.3 & 36.2 \\
\textbf{R} & 55.9 & 46.6 & 19.9 & 10.7 & 33.3 \\
\textbf{P} & 48.9 & 51.0 & 51.8 & 26.5 & 44.5 \\
\bottomrule
\end{tabular}
}
\caption{End-to-end performance on LORELEI (under-labeled for end-to-end linking).
}
\label{Results-lorelei}
\end{table}

\begin{table}[h!]
\small
\centering
{
\begin{tabular}{lccc} \hline
Model & & TAC-KBP\\ \hline
& es & zh & macro-avg \\ \hline
\citet{tsai2016cross} & 82.4 & 85.1 & 83.8  \\
\citet{sil-et-al},  &  83.9 & 85.9 & 84.9  \\
\citet{upadhyay-etal-2018-joint} & 84.4 & 86.0& 85.2  \\
\citet{zhou-etal-2019-towards} &  82.9 & 85.5 & 84.2 \\
\midrule
\citet{de-cao-etal-2022-multilingual} & 86.3 & 64.6& 75.5
\\
\textbf{BELA} & 88.7 & 75.3 & 82.0 \\
\bottomrule
\end{tabular}
}
\caption{ED accuracy on TAC-KBP. BELA outperforms baseline models on es and is competitive on zh.}
\label{Results-TAC-KBP}
\end{table}

We report ED results on Mewsli-9 in Table \ref{Results-ED}. Note that to ensure comparability, here we use original Mewsli-9 data and a model that was not finetuned on any Mewsli-9 data. BELA falls short of the current state-of-the art multilingual entity disambiguation model mGENRE by an average of 3\% points.

Note that mGENRE model is a sequence-to-sequence model that was specifically trained for ED. It requires one pass of encoding for each mention and can rely on special entity delimiter tokens. BELA is trained end-to-end without mention delimiters, disambiguating multiple entities in a single pass. Also, mGENRE's underlying language model was trained for 500k more steps on language modeling before fine-tuning for entity disambiguation. Finally, mGENRE is computationally more expensive at inference making it impractical for real world applications, see Section~\ref{complexity} for more details. BELA is only slightly less accurate than \citet{botha-etal-2020-entity}, which however requires, like mGENRE, a mention encoding pass for each candidate mention.

Results on TAC-KBP are reported in Table \ref{Results-TAC-KBP}. Note that all the baselines except \citet{de-cao-etal-2022-multilingual} utilize a predefined candidate set for each mention (also referred as mention table) for entity disambiguation. Whereas, BELA and \citet{de-cao-etal-2022-multilingual} do not require such candidate set. For \textit{es}, BELA outperforms all baseline models (including the ones that use mention table). For \textit{zh}, BELA outperforms \citet{de-cao-etal-2022-multilingual} and is competitive with other baselines that rely on the candidate set.

\subsection{Rejection head}
\begin{figure}[ht!]
\centering
     \includegraphics[clip,width=0.45\textwidth]{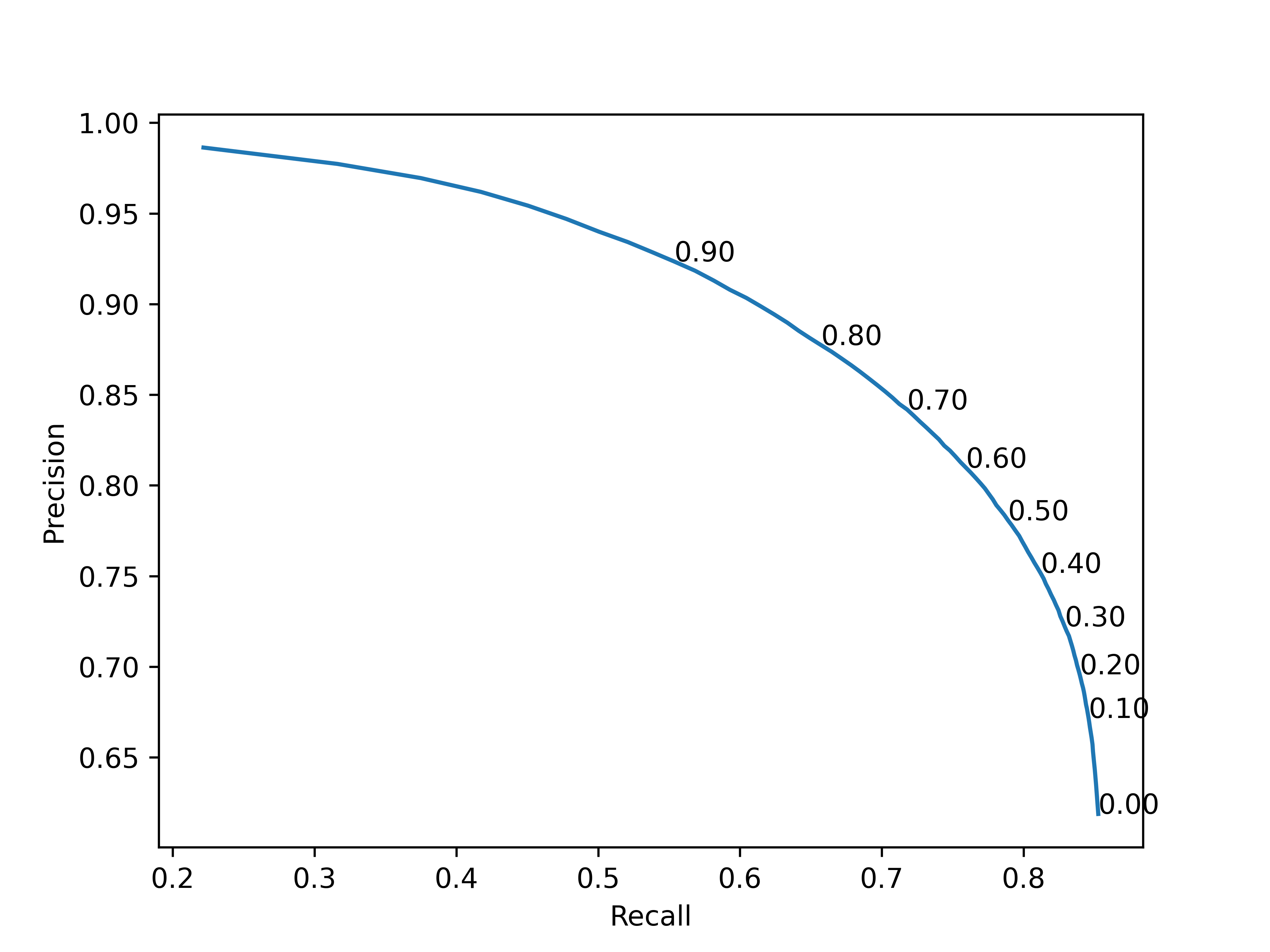}
\caption{Precision and recall curve for Mewsli-9'-test derived using the rejection head. The values written on the curve are the corresponding $\gamma$ thresholds.
\label{fig:reject-head}}

\end{figure}

In Figure \ref{fig:reject-head}, we show how we can vary hyper-parameter $\gamma$ (ED score threshold) to regulate between precision and recall. This is important as applications vary in the requirements they impose on either metrics.

\subsection{Computational Complexity}
\label{complexity}
We report on BELA's computational complexity at inference time: for a discussion of the computational requirements for training, see Appendix \ref{train_complex}. We show that BELA's fast inference time makes it suitable for integration in large scale NLU systems. All numbers are reported using 40GB A100 GPUs.

We compare the \textit{end-to-end} inference speed of BELA with that of the monolingual end-to-end GENRE, which is architecturally close to the current state-of-the art multilingual ED model mGENRE. We computed inference time on the English subset of Mewsli-9' and report number of samples per second on GPU feeding the models with one input at a time (i.e., batch size of 1). We report results in Table \ref{tab:el-speed}. We find that BELA is more than 50 times faster than monolingual end-to-end GENRE. Increasing the batch size to 1024, BELA hits a throughput of 53 samples per second. 

Using the same dataset, we also compare BELA's \textit{disambiguation}. We compare with \citet{decao2021autoregressive}  and \citet{Wu2020ScalableZE} and show that our model outperforms other models by a large margin (see Table \ref{tab:ed-speed}).

\begin{table}[h!]
\centering
\small
\begin{tabular}{lc}
\toprule
\textbf{Method} & \textbf{\# Samples / Sec}  \\
\midrule
\citet{decao2021autoregressive}             & 0.22\\
\midrule
BELA                                        & \textbf{11.11}\\
\bottomrule
\end{tabular}
\caption{Measured inference speed of BELA vs. monolingual GENRE. BELA is $\sim$ 50 times faster.}
\label{tab:el-speed}
\end{table}

\begin{table}[h!]
\centering
\small
\begin{tabular}{lc}
\toprule
\textbf{Method} & \textbf{\# Mentions / Sec}  \\
\midrule
\citet{decao2021autoregressive}             & 0.44\\
\midrule
\citet{Wu2020ScalableZE}              & 13\\
\midrule
BELA                                        & \textbf{114}\\
\bottomrule
\end{tabular}
\caption{Measured disambiguation speed.}
\label{tab:ed-speed}
\end{table}

\section{Related Work}
EL is an extensively studied task. Prior to the introduction of large language models (LLMs), EL systems used frequency and typing information, alias tables, TF-IDF-based methods
and neural networks to model context, mention
and entity \cite{cucerzan-2007-large, bunescu-pasca-2006-using, Milne2008LearningTL, he-etal-2013-learning, Sun2015ModelingMC, lazic-etal-2015-plato, Raiman2018DeepTypeME, kolitsas-etal-2018-end, gupta-etal-2017-entity, ganea-hofmann-2017-deep, DBLP:journals/corr/abs-1811-10547, DBLP:journals/corr/abs-1909-05780}.
With the introduction of LLMs, EL sytems shifted to transformer-based architectures. \citet{gillick-etal-2019-learning} present a dual encoder architecture based on BERT \cite{devlin-etal-2019-bert} that encodes mentions and entities in
the same dense vector space and performs EL via kNN search. \citet{logeswaran-etal-2019-zero} proposed zero-short EL and showed that domain adaptive training can address the domain shift problem. Subsequently, \citet{Wu2020ScalableZE} showed that pre-trained zero-shot architectures are both highly accurate and computationally efficient at scale. \citet{li2020elq} introduced a single-pass end-to-end architectures that proved successful on short text spans. Similarly, \citet{ayoola-etal-2022-refined} also relies on an efficient single-pass architecture that also incorporates type information for end-to-end linking. All these works target English only data.

\citet{botha-etal-2020-entity} extended these architecture to a \textit{multilingual setting} enabling EL of mentions in more than 100 languages to a language agnostic reference KB. In contrast to their work, we introduce an end-to-end model that jointly detects and disambiguates mentions in a multilingual setting. 

Before that, EL work outside of the standard monolingual setting focused on a \textit{cross-lingual setting} \cite{McNamee2011CrossLanguageEL, tsai-roth-2016-cross, sil-etal-2018-multi} where mentions in one language have to link to a reference KB expressed in another language typically limited to English Wikipedia. Work in this direction features progress on low-resource languages \cite{zhou-etal-2020-improving-candidate}, zero-shot transfer \cite{sil-florian-2016-one, Rijhwani2018ZeroshotNT, zhou-etal-2019-towards} and scaling to many languages \cite{pan-etal-2017-cross}. Note that in this work we focus on the more general multilingual setting where we can link to entities that might not be represented in a target monolingual KB.

Besides bi-encoder architectures, \citet{decao2021autoregressive} introduce a model that links entities by generating unique tags in an autoregressive fashion. In mGENRE \cite{de-cao-etal-2022-multilingual}, they extended this approach to a multilingual setting. mGENRE relies on gold mentions marked by special entity delimiter tokens.

In follow-up work they addressed the efficiency problems of generative EL by parallelizing autoregressive linking across mentions and relying on shallow and efficient decoding \cite{de-cao-etal-2021-highly}. This work is not multilingual and relays on a pre-computed set of candidates. Generative EL methods are as for now less established and impractical for real world applications.

\section{Conclusions}

We have presented BELA, the first end-to-end entity linking system that covers 97 languages and is suitable for large-scale applications. BELA scales to 16 million entities with a measured throughput of 53 samples per second on a single GPU.

We release BELA’s codebase and model in the hope that it will find its way in useful applications.

\section*{Ethical Considerations}

It is important to consider the ethical implications of EL models, especially because EL has many real-world applications.

EL can be deployed both in non-problematic situations, such as content analysis for understanding user tastes or detecting hate speech, but also in more controversial applications, such as surveillance.

EL models can also be biased.
BELA is a multilingual model that relies on Wikidata as its knowledge base, which is itself based on Wikipedia articles.
However Wikipedia suffers from well-known biases related to geographical location \cite{kaffee-etal-2018-learning,beytia2020positioning}, gender \cite{hinnosaar2019gender,schmahl-etal-2020-wikipedia}, and marginalized groups \cite{worku2020exploring}.
Having more articles for certain languages or groups can create allocation bias in downstream applications, meaning that the model will perform better for the high-resource groups with more Wikipedia articles.

Thanks to the dense retrieval architecture, the model's index can however be updated hand in hand with efforts to reduce biases in Wikipedia\footnote{\url{https://en.wikipedia.org/wiki/Wikipedia:WikiProject_Women}},\footnote{\url{https://en.wikipedia.org/wiki/Wikipedia:WikiProject_Women_in_Red}}, hence mitigating biases in the model.
By releasing this first end-to-end multilingual EL system, we also open the way for using EL not only for English but for a wider range of languages, to the benefit of a wider audience over the world.
We hope that releasing our code and model will improve reproducibility and foster open-science for the EL community.

\bibliography{anthology,bibliography}

\begin{thebibliography}{51}
\expandafter\ifx\csname natexlab\endcsname\relax\def\natexlab#1{#1}\fi

\bibitem[{Ayoola et~al.(2022)Ayoola, Tyagi, Fisher, Christodoulopoulos, and
  Pierleoni}]{ayoola-etal-2022-refined}
Tom Ayoola, Shubhi Tyagi, Joseph Fisher, Christos Christodoulopoulos, and
  Andrea Pierleoni. 2022.
\newblock \href {https://doi.org/10.18653/v1/2022.naacl-industry.24}
  {{R}e{F}in{ED}: An efficient zero-shot-capable approach to end-to-end entity
  linking}.
\newblock In \emph{Proceedings of the 2022 Conference of the North American
  Chapter of the Association for Computational Linguistics: Human Language
  Technologies: Industry Track}, pages 209--220, Hybrid: Seattle, Washington +
  Online. Association for Computational Linguistics.

\bibitem[{Beyt{\'\i}a(2020)}]{beytia2020positioning}
Pablo Beyt{\'\i}a. 2020.
\newblock The positioning matters: Estimating geographical bias in the
  multilingual record of biographies on wikipedia.
\newblock In \emph{Companion Proceedings of the Web Conference 2020}, pages
  806--810.

\bibitem[{Botha et~al.(2020)Botha, Shan, and Gillick}]{botha-etal-2020-entity}
Jan~A. Botha, Zifei Shan, and Daniel Gillick. 2020.
\newblock \href {https://doi.org/10.18653/v1/2020.emnlp-main.630} {{E}ntity
  {L}inking in 100 {L}anguages}.
\newblock In \emph{Proceedings of the 2020 Conference on Empirical Methods in
  Natural Language Processing (EMNLP)}, pages 7833--7845, Online. Association
  for Computational Linguistics.

\bibitem[{Broscheit(2019)}]{broscheit-2019-investigating}
Samuel Broscheit. 2019.
\newblock \href {https://doi.org/10.18653/v1/K19-1063} {Investigating entity
  knowledge in {BERT} with simple neural end-to-end entity linking}.
\newblock In \emph{Proceedings of the 23rd Conference on Computational Natural
  Language Learning (CoNLL)}, pages 677--685, Hong Kong, China. Association for
  Computational Linguistics.

\bibitem[{Bunescu and Pa{\c{s}}ca(2006)}]{bunescu-pasca-2006-using}
Razvan Bunescu and Marius Pa{\c{s}}ca. 2006.
\newblock \href {https://www.aclweb.org/anthology/E06-1002} {Using encyclopedic
  knowledge for named entity disambiguation}.
\newblock In \emph{11th Conference of the {E}uropean Chapter of the Association
  for Computational Linguistics}, Trento, Italy. Association for Computational
  Linguistics.

\bibitem[{Cucerzan(2007)}]{cucerzan-2007-large}
Silviu Cucerzan. 2007.
\newblock \href {https://www.aclweb.org/anthology/D07-1074} {Large-scale named
  entity disambiguation based on {W}ikipedia data}.
\newblock In \emph{Proceedings of the 2007 Joint Conference on Empirical
  Methods in Natural Language Processing and Computational Natural Language
  Learning ({EMNLP}-{C}o{NLL})}, pages 708--716, Prague, Czech Republic.
  Association for Computational Linguistics.

\bibitem[{De~Cao et~al.(2021)De~Cao, Aziz, and Titov}]{de-cao-etal-2021-highly}
Nicola De~Cao, Wilker Aziz, and Ivan Titov. 2021.
\newblock \href {https://doi.org/10.18653/v1/2021.emnlp-main.604} {Highly
  parallel autoregressive entity linking with discriminative correction}.
\newblock In \emph{Proceedings of the 2021 Conference on Empirical Methods in
  Natural Language Processing}, pages 7662--7669, Online and Punta Cana,
  Dominican Republic. Association for Computational Linguistics.

\bibitem[{{De Cao} et~al.(2021){De Cao}, Izacard, Riedel, and
  Petroni}]{decao2021autoregressive}
Nicola {De Cao}, Gautier Izacard, Sebastian Riedel, and Fabio Petroni. 2021.
\newblock \href {https://openreview.net/forum?id=5k8F6UU39V} {Autoregressive
  entity retrieval}.
\newblock In \emph{9th International Conference on Learning Representations,
  {ICLR} 2021, Virtual Event, Austria, May 3-7, 2021}. OpenReview.net.

\bibitem[{De~Cao et~al.(2022)De~Cao, Wu, Popat, Artetxe, Goyal, Plekhanov,
  Zettlemoyer, Cancedda, Riedel, and Petroni}]{de-cao-etal-2022-multilingual}
Nicola De~Cao, Ledell Wu, Kashyap Popat, Mikel Artetxe, Naman Goyal, Mikhail
  Plekhanov, Luke Zettlemoyer, Nicola Cancedda, Sebastian Riedel, and Fabio
  Petroni. 2022.
\newblock \href {https://doi.org/10.1162/tacl_a_00460} {Multilingual
  autoregressive entity linking}.
\newblock \emph{Transactions of the Association for Computational Linguistics},
  10:274--290.

\bibitem[{Devlin et~al.(2019)Devlin, Chang, Lee, and
  Toutanova}]{devlin-etal-2019-bert}
Jacob Devlin, Ming-Wei Chang, Kenton Lee, and Kristina Toutanova. 2019.
\newblock \href {https://doi.org/10.18653/v1/N19-1423} {{BERT}: Pre-training of
  deep bidirectional transformers for language understanding}.
\newblock In \emph{Proceedings of the 2019 Conference of the North {A}merican
  Chapter of the Association for Computational Linguistics: Human Language
  Technologies, Volume 1 (Long and Short Papers)}, pages 4171--4186,
  Minneapolis, Minnesota. Association for Computational Linguistics.

\bibitem[{F{\'e}vry et~al.(2020)F{\'e}vry, FitzGerald, Soares, and
  Kwiatkowski}]{fevry2020empirical}
Thibault F{\'e}vry, Nicholas FitzGerald, Livio~Baldini Soares, and Tom
  Kwiatkowski. 2020.
\newblock \href {https://openreview.net/forum?id=iHXV8UGYyL} {Empirical
  evaluation of pretraining strategies for supervised entity linking}.
\newblock In \emph{Automated Knowledge Base Construction (AKAB)}.

\bibitem[{Ganea and Hofmann(2017)}]{ganea-hofmann-2017-deep}
Octavian-Eugen Ganea and Thomas Hofmann. 2017.
\newblock \href {https://doi.org/10.18653/v1/D17-1277} {Deep joint entity
  disambiguation with local neural attention}.
\newblock In \emph{Proceedings of the 2017 Conference on Empirical Methods in
  Natural Language Processing}, pages 2619--2629, Copenhagen, Denmark.
  Association for Computational Linguistics.

\bibitem[{Gillick et~al.(2019)Gillick, Kulkarni, Lansing, Presta, Baldridge,
  Ie, and Garcia-Olano}]{gillick-etal-2019-learning}
Daniel Gillick, Sayali Kulkarni, Larry Lansing, Alessandro Presta, Jason
  Baldridge, Eugene Ie, and Diego Garcia-Olano. 2019.
\newblock \href {https://doi.org/10.18653/v1/K19-1049} {Learning dense
  representations for entity retrieval}.
\newblock In \emph{Proceedings of the 23rd Conference on Computational Natural
  Language Learning (CoNLL)}, pages 528--537, Hong Kong, China. Association for
  Computational Linguistics.

\bibitem[{Gupta et~al.(2017)Gupta, Singh, and Roth}]{gupta-etal-2017-entity}
Nitish Gupta, Sameer Singh, and Dan Roth. 2017.
\newblock \href {https://doi.org/10.18653/v1/D17-1284} {Entity linking via
  joint encoding of types, descriptions, and context}.
\newblock In \emph{Proceedings of the 2017 Conference on Empirical Methods in
  Natural Language Processing}, pages 2681--2690, Copenhagen, Denmark.
  Association for Computational Linguistics.

\bibitem[{He et~al.(2013)He, Liu, Li, Zhou, Zhang, and
  Wang}]{he-etal-2013-learning}
Zhengyan He, Shujie Liu, Mu~Li, Ming Zhou, Longkai Zhang, and Houfeng Wang.
  2013.
\newblock \href {https://www.aclweb.org/anthology/P13-2006} {Learning entity
  representation for entity disambiguation}.
\newblock In \emph{Proceedings of the 51st Annual Meeting of the Association
  for Computational Linguistics (Volume 2: Short Papers)}, pages 30--34, Sofia,
  Bulgaria. Association for Computational Linguistics.

\bibitem[{Hinnosaar(2019)}]{hinnosaar2019gender}
Marit Hinnosaar. 2019.
\newblock Gender inequality in new media: Evidence from wikipedia.
\newblock \emph{Journal of economic behavior \& organization}, 163:262--276.

\bibitem[{Hoffart et~al.(2011)Hoffart, Yosef, Bordino, F{\"u}rstenau, Pinkal,
  Spaniol, Taneva, Thater, and Weikum}]{hoffart-etal-2011-robust}
Johannes Hoffart, Mohamed~Amir Yosef, Ilaria Bordino, Hagen F{\"u}rstenau,
  Manfred Pinkal, Marc Spaniol, Bilyana Taneva, Stefan Thater, and Gerhard
  Weikum. 2011.
\newblock \href {https://www.aclweb.org/anthology/D11-1072} {Robust
  disambiguation of named entities in text}.
\newblock In \emph{Proceedings of the 2011 Conference on Empirical Methods in
  Natural Language Processing}, pages 782--792, Edinburgh, Scotland, UK.
  Association for Computational Linguistics.

\bibitem[{Ji et~al.(2015)Ji, Nothman, Hachey, and Florian}]{ji2015overview}
Heng Ji, Joel Nothman, Ben Hachey, and Radu Florian. 2015.
\newblock Overview of tac-kbp2015 tri-lingual entity discovery and linking.
\newblock In \emph{TAC}.

\bibitem[{Johnson et~al.(2017)Johnson, Douze, and J{\'e}gou}]{JDH17}
Jeff Johnson, Matthijs Douze, and Herv{\'e} J{\'e}gou. 2017.
\newblock Billion-scale similarity search with gpus.
\newblock \emph{arXiv preprint arXiv:1702.08734}.

\bibitem[{Kaffee et~al.(2018)Kaffee, Elsahar, Vougiouklis, Gravier, Laforest,
  Hare, and Simperl}]{kaffee-etal-2018-learning}
Lucie-Aim{\'e}e Kaffee, Hady Elsahar, Pavlos Vougiouklis, Christophe Gravier,
  Fr{\'e}d{\'e}rique Laforest, Jonathon Hare, and Elena Simperl. 2018.
\newblock \href {https://doi.org/10.18653/v1/N18-2101} {Learning to generate
  {W}ikipedia summaries for underserved languages from {W}ikidata}.
\newblock In \emph{Proceedings of the 2018 Conference of the North {A}merican
  Chapter of the Association for Computational Linguistics: Human Language
  Technologies, Volume 2 (Short Papers)}, pages 640--645, New Orleans,
  Louisiana. Association for Computational Linguistics.

\bibitem[{Khalife and Vazirgiannis(2018)}]{DBLP:journals/corr/abs-1811-10547}
Sammy Khalife and Michalis Vazirgiannis. 2018.
\newblock \href {http://arxiv.org/abs/1811.10547} {Scalable graph-based
  individual named entity identification}.
\newblock \emph{CoRR}, abs/1811.10547.

\bibitem[{Kolitsas et~al.(2018)Kolitsas, Ganea, and
  Hofmann}]{kolitsas-etal-2018-end}
Nikolaos Kolitsas, Octavian-Eugen Ganea, and Thomas Hofmann. 2018.
\newblock \href {https://doi.org/10.18653/v1/K18-1050} {End-to-end neural
  entity linking}.
\newblock In \emph{Proceedings of the 22nd Conference on Computational Natural
  Language Learning}, pages 519--529, Brussels, Belgium. Association for
  Computational Linguistics.

\bibitem[{Lample and Conneau(2019)}]{lample2019cross}
Guillaume Lample and Alexis Conneau. 2019.
\newblock Cross-lingual language model pretraining.
\newblock \emph{Advances in Neural Information Processing Systems (NeurIPS)}.

\bibitem[{Lazic et~al.(2015)Lazic, Subramanya, Ringgaard, and
  Pereira}]{lazic-etal-2015-plato}
Nevena Lazic, Amarnag Subramanya, Michael Ringgaard, and Fernando Pereira.
  2015.
\newblock \href {https://doi.org/10.1162/tacl_a_00154} {{P}lato: A selective
  context model for entity resolution}.
\newblock \emph{Transactions of the Association for Computational Linguistics},
  3:503--515.

\bibitem[{Li et~al.(2020)Li, Min, Iyer, Mehdad, and Yih}]{li2020elq}
Belinda~Z Li, Sewon Min, Srinivasan Iyer, Yashar Mehdad, and Wen-tau Yih. 2020.
\newblock \href {https://aclanthology.org/2020.emnlp-main.522.pdf} {Efficient
  one-pass end-to-end entity linking for questions}.
\newblock In \emph{Proceedings of the Conference on Empirical Methods in
  Natural Language Processing (EMNLP)}.

\bibitem[{Logeswaran et~al.(2019)Logeswaran, Chang, Lee, Toutanova, Devlin, and
  Lee}]{logeswaran-etal-2019-zero}
Lajanugen Logeswaran, Ming-Wei Chang, Kenton Lee, Kristina Toutanova, Jacob
  Devlin, and Honglak Lee. 2019.
\newblock \href {https://doi.org/10.18653/v1/P19-1335} {Zero-shot entity
  linking by reading entity descriptions}.
\newblock In \emph{Proceedings of the 57th Annual Meeting of the Association
  for Computational Linguistics}, pages 3449--3460, Florence, Italy.
  Association for Computational Linguistics.

\bibitem[{Martins et~al.(2019)Martins, Marinho, and
  Martins}]{martins-etal-2019-joint}
Pedro~Henrique Martins, Zita Marinho, and Andr{\'e} F.~T. Martins. 2019.
\newblock \href {https://doi.org/10.18653/v1/P19-2026} {Joint learning of named
  entity recognition and entity linking}.
\newblock In \emph{Proceedings of the 57th Annual Meeting of the Association
  for Computational Linguistics: Student Research Workshop}, pages 190--196,
  Florence, Italy. Association for Computational Linguistics.

\bibitem[{McNamee et~al.(2011)McNamee, Mayfield, Lawrie, Oard, and
  Doermann}]{McNamee2011CrossLanguageEL}
Paul McNamee, James Mayfield, Dawn~J Lawrie, Douglas~W. Oard, and David~S.
  Doermann. 2011.
\newblock Cross-language entity linking.
\newblock In \emph{International Joint Conference on Natural Language
  Processing}.

\bibitem[{Milne and Witten(2008)}]{Milne2008LearningTL}
David~N. Milne and Ian~H. Witten. 2008.
\newblock Learning to link with wikipedia.
\newblock In \emph{CIKM '08}.

\bibitem[{Onoe and Durrett(2019)}]{DBLP:journals/corr/abs-1909-05780}
Yasumasa Onoe and Greg Durrett. 2019.
\newblock \href {http://arxiv.org/abs/1909.05780} {Fine-grained entity typing
  for domain independent entity linking}.
\newblock \emph{CoRR}, abs/1909.05780.

\bibitem[{Pan et~al.(2017)Pan, Zhang, May, Nothman, Knight, and
  Ji}]{pan-etal-2017-cross}
Xiaoman Pan, Boliang Zhang, Jonathan May, Joel Nothman, Kevin Knight, and Heng
  Ji. 2017.
\newblock \href {https://doi.org/10.18653/v1/P17-1178} {Cross-lingual name
  tagging and linking for 282 languages}.
\newblock In \emph{Proceedings of the 55th Annual Meeting of the Association
  for Computational Linguistics (Volume 1: Long Papers)}, pages 1946--1958,
  Vancouver, Canada. Association for Computational Linguistics.

\bibitem[{Peters et~al.(2019)Peters, Neumann, Logan, Schwartz, Joshi, Singh,
  and Smith}]{peters-etal-2019-knowledge}
Matthew~E. Peters, Mark Neumann, Robert Logan, Roy Schwartz, Vidur Joshi,
  Sameer Singh, and Noah~A. Smith. 2019.
\newblock \href {https://doi.org/10.18653/v1/D19-1005} {Knowledge enhanced
  contextual word representations}.
\newblock In \emph{Proceedings of the 2019 Conference on Empirical Methods in
  Natural Language Processing and the 9th International Joint Conference on
  Natural Language Processing (EMNLP-IJCNLP)}, pages 43--54, Hong Kong, China.
  Association for Computational Linguistics.

\bibitem[{Raiman and Raiman(2018)}]{Raiman2018DeepTypeME}
Jonathan Raiman and Olivier Raiman. 2018.
\newblock Deeptype: Multilingual entity linking by neural type system
  evolution.
\newblock In \emph{AAAI}.

\bibitem[{Rijhwani et~al.(2018)Rijhwani, Xie, Neubig, and
  Carbonell}]{Rijhwani2018ZeroshotNT}
Shruti Rijhwani, Jiateng Xie, Graham Neubig, and Jaime~G. Carbonell. 2018.
\newblock Zero-shot neural transfer for cross-lingual entity linking.
\newblock In \emph{AAAI Conference on Artificial Intelligence}.

\bibitem[{Schmahl et~al.(2020)Schmahl, Viering, Makrodimitris, Naseri~Jahfari,
  Tax, and Loog}]{schmahl-etal-2020-wikipedia}
Katja~Geertruida Schmahl, Tom~Julian Viering, Stavros Makrodimitris, Arman
  Naseri~Jahfari, David Tax, and Marco Loog. 2020.
\newblock \href {https://doi.org/10.18653/v1/2020.nlpcss-1.11} {Is {W}ikipedia
  succeeding in reducing gender bias? assessing changes in gender bias in
  {W}ikipedia using word embeddings}.
\newblock In \emph{Proceedings of the Fourth Workshop on Natural Language
  Processing and Computational Social Science}, pages 94--103, Online.
  Association for Computational Linguistics.

\bibitem[{Shen et~al.(2015)Shen, Wang, and Han}]{6823700}
Wei Shen, Jianyong Wang, and Jiawei Han. 2015.
\newblock \href {https://doi.org/10.1109/TKDE.2014.2327028} {Entity linking
  with a knowledge base: Issues, techniques, and solutions}.
\newblock \emph{IEEE Transactions on Knowledge and Data Engineering},
  27(2):443--460.

\bibitem[{Sil et~al.(2018{\natexlab{a}})Sil, Ji, Roth, and
  Cucerzan}]{sil-etal-2018-multi}
Avi Sil, Heng Ji, Dan Roth, and Silviu-Petru Cucerzan. 2018{\natexlab{a}}.
\newblock \href {https://doi.org/10.18653/v1/P18-5008} {Multi-lingual entity
  discovery and linking}.
\newblock In \emph{Proceedings of the 56th Annual Meeting of the Association
  for Computational Linguistics: Tutorial Abstracts}, pages 22--29, Melbourne,
  Australia. Association for Computational Linguistics.

\bibitem[{Sil and Florian(2016)}]{sil-florian-2016-one}
Avirup Sil and Radu Florian. 2016.
\newblock \href {https://doi.org/10.18653/v1/P16-1213} {One for all: Towards
  language independent named entity linking}.
\newblock In \emph{Proceedings of the 54th Annual Meeting of the Association
  for Computational Linguistics (Volume 1: Long Papers)}, pages 2255--2264,
  Berlin, Germany. Association for Computational Linguistics.

\bibitem[{Sil et~al.(2018{\natexlab{b}})Sil, Kundu, Florian, and
  Hamza}]{sil-et-al}
Avirup Sil, Gourab Kundu, Radu Florian, and Wael Hamza. 2018{\natexlab{b}}.
\newblock Neural cross-lingual entity linking.
\newblock In \emph{Proceedings of the Thirty-Second AAAI Conference on
  Artificial Intelligence and Thirtieth Innovative Applications of Artificial
  Intelligence Conference and Eighth AAAI Symposium on Educational Advances in
  Artificial Intelligence}. AAAI Press.

\bibitem[{Strassel and Tracey(2016)}]{strassel-tracey-2016-lorelei}
Stephanie Strassel and Jennifer Tracey. 2016.
\newblock \href {https://www.aclweb.org/anthology/L16-1521} {{LORELEI} language
  packs: Data, tools, and resources for technology development in low resource
  languages}.
\newblock In \emph{Proceedings of the Tenth International Conference on
  Language Resources and Evaluation ({LREC} 2016)}, pages 3273--3280,
  Portoro{\v{z}}, Slovenia. European Language Resources Association (ELRA).

\bibitem[{Sun et~al.(2015)Sun, Lin, Tang, Yang, Ji, and
  Wang}]{Sun2015ModelingMC}
Yaming Sun, Lei Lin, Duyu Tang, Nan Yang, Zhenzhou Ji, and Xiaolong Wang. 2015.
\newblock Modeling mention, context and entity with neural networks for entity
  disambiguation.
\newblock In \emph{IJCAI}.

\bibitem[{Tsai and Roth(2016{\natexlab{a}})}]{tsai2016cross}
Chen-Tse Tsai and Dan Roth. 2016{\natexlab{a}}.
\newblock Cross-lingual wikification using multilingual embeddings.
\newblock In \emph{Proceedings of the 2016 Conference of the North American
  Chapter of the Association for Computational Linguistics: Human Language
  Technologies}, pages 589--598.

\bibitem[{Tsai and Roth(2016{\natexlab{b}})}]{tsai-roth-2016-cross}
Chen-Tse Tsai and Dan Roth. 2016{\natexlab{b}}.
\newblock \href {https://doi.org/10.18653/v1/N16-1072} {Cross-lingual
  wikification using multilingual embeddings}.
\newblock In \emph{Proceedings of the 2016 Conference of the North {A}merican
  Chapter of the Association for Computational Linguistics: Human Language
  Technologies}, pages 589--598, San Diego, California. Association for
  Computational Linguistics.

\bibitem[{Upadhyay et~al.(2018)Upadhyay, Gupta, and
  Roth}]{upadhyay-etal-2018-joint}
Shyam Upadhyay, Nitish Gupta, and Dan Roth. 2018.
\newblock \href {https://doi.org/10.18653/v1/D18-1270} {Joint multilingual
  supervision for cross-lingual entity linking}.
\newblock In \emph{Proceedings of the 2018 Conference on Empirical Methods in
  Natural Language Processing}, pages 2486--2495, Brussels, Belgium.
  Association for Computational Linguistics.

\bibitem[{Vrande{\v{c}}i{\'c} and Kr{\"o}tzsch(2014)}]{vrandevcic2014wikidata}
Denny Vrande{\v{c}}i{\'c} and Markus Kr{\"o}tzsch. 2014.
\newblock Wikidata: a free collaborative knowledgebase.
\newblock \emph{Communications of the ACM}, 57(10):78--85.

\bibitem[{Worku et~al.(2020)Worku, Bipat, McDonald, and
  Zachry}]{worku2020exploring}
Zena Worku, Taryn Bipat, David~W. McDonald, and Mark Zachry. 2020.
\newblock \href {https://doi.org/10.1145/3412569.3412573} {Exploring systematic
  bias through article deletions on wikipedia from a behavioral perspective}.
\newblock In \emph{Proceedings of the 16th International Symposium on Open
  Collaboration}, OpenSym '20, New York, NY, USA. Association for Computing
  Machinery.

\bibitem[{Wu et~al.(2020{\natexlab{a}})Wu, Petroni, Josifoski, Riedel, and
  Zettlemoyer}]{wu-etal-2020-scalable}
Ledell Wu, Fabio Petroni, Martin Josifoski, Sebastian Riedel, and Luke
  Zettlemoyer. 2020{\natexlab{a}}.
\newblock \href {https://doi.org/10.18653/v1/2020.emnlp-main.519} {Scalable
  zero-shot entity linking with dense entity retrieval}.
\newblock In \emph{Proceedings of the 2020 Conference on Empirical Methods in
  Natural Language Processing (EMNLP)}, pages 6397--6407, Online. Association
  for Computational Linguistics.

\bibitem[{Wu et~al.(2020{\natexlab{b}})Wu, Petroni, Josifoski, Riedel, and
  Zettlemoyer}]{Wu2020ScalableZE}
Ledell~Yu Wu, Fabio Petroni, Martin Josifoski, Sebastian Riedel, and Luke
  Zettlemoyer. 2020{\natexlab{b}}.
\newblock Scalable zero-shot entity linking with dense entity retrieval.
\newblock In \emph{EMNLP}.

\bibitem[{Yin et~al.(2016)Yin, Yu, Xiang, Zhou, and
  Sch{\"u}tze}]{yin-etal-2016-simple}
Wenpeng Yin, Mo~Yu, Bing Xiang, Bowen Zhou, and Hinrich Sch{\"u}tze. 2016.
\newblock \href {https://www.aclweb.org/anthology/C16-1164} {Simple question
  answering by attentive convolutional neural network}.
\newblock In \emph{Proceedings of {COLING} 2016, the 26th International
  Conference on Computational Linguistics: Technical Papers}, pages 1746--1756,
  Osaka, Japan. The COLING 2016 Organizing Committee.

\bibitem[{Zhou et~al.(2019)Zhou, Rijhwani, and Neubig}]{zhou-etal-2019-towards}
Shuyan Zhou, Shruti Rijhwani, and Graham Neubig. 2019.
\newblock \href {https://doi.org/10.18653/v1/D19-6127} {Towards zero-resource
  cross-lingual entity linking}.
\newblock In \emph{Proceedings of the 2nd Workshop on Deep Learning Approaches
  for Low-Resource NLP (DeepLo 2019)}, pages 243--252, Hong Kong, China.
  Association for Computational Linguistics.

\bibitem[{Zhou et~al.(2020)Zhou, Rijhwani, Wieting, Carbonell, and
  Neubig}]{zhou-etal-2020-improving-candidate}
Shuyan Zhou, Shruti Rijhwani, John Wieting, Jaime Carbonell, and Graham Neubig.
  2020.
\newblock \href {https://doi.org/10.1162/tacl_a_00303} {Improving candidate
  generation for low-resource cross-lingual entity linking}.
\newblock \emph{Transactions of the Association for Computational Linguistics},
  8:109--124.

\end{thebibliography}
\bibliographystyle{acl_natbib}

\clearpage
\appendix
\section{Hyper-parameters}

The mention detection threshold  $\beta$ and entity linking threshold  $\gamma$ are tuned on the AIDA development optimizing for F1. We report their value in Tables \ref{tab:hyperparameters1}, \ref{tab:hyperparameters2}.

\label{hyperparameters}
\begin{table}[t]
\centering
\small
\begin{tabular}{lc}
\toprule
\textbf{Hyper-parameter} & \textbf{Value} \\
\midrule
$\gamma$ & 0.2\\
$\beta$ & 0.4 \\
\bottomrule
\end{tabular}
\caption{Hyper-parameters for model finetuned on Mewsli-9'.}
\label{tab:hyperparameters1}
\end{table}

\label{hyperparameters}
\begin{table}[t]
\centering
\small
\begin{tabular}{lc}
\toprule
\textbf{Hyper-parameter} & \textbf{Value} \\
\midrule
$\gamma$ & 0.4\\
$\beta$ & 0.0 \\
\bottomrule
\end{tabular}
\caption{Hyper-parameters for model finetuned on AIDA.}
\label{tab:hyperparameters2}
\end{table}

\begin{table*}[h!]
    \centering
    \small
    \begin{tabular}{cc|cc|cc|cc}
        \toprule
         \textbf{Code} & \textbf{Name} & \textbf{Code} & \textbf{Name} & \textbf{Code} & \textbf{Name} & \textbf{Code} & \textbf{Name} \\
         \midrule
        \textit{af} & Afrikaans & \textit{als} & Alemannic & \textit{am} & Amharic & \textit{an} &  Aragonese \\ 
        \textit{ang} & Old English & \textit{ar} & Arabic & \textit{arz} & Egyptian Arabic & \textit{ast} & Asturian \\ 
        \textit{az} & Azerbaijani & \textit{bar} & Bavarian & \textit{be} & Belarusian & \textit{bg} & Bulgarian \\ 
        \textit{bn} & Bengali & \textit{br} & Breton & \textit{bs} & Bosnian & \textit{ca} & Catalan \\ 
         \textit{ckb} & Central Kurdish & \textit{cs} & Czech & \textit{cy} & Welsh & \textit{da} & Danish \\ 
         \textit{de} & German & \textit{el} & Greek & \textit{en} & English & \textit{eo} & Esperanto \\ 
         \textit{es} & Spanish & \textit{et} & Estonian & \textit{eu} & Basque & \textit{fa} & Persian \\ 
        \textit{fi} & Finnish & \textit{fr} & French & \textit{fy} & Western Frisian & \textit{ga} & Irish \\ 
        \textit{gan} & Gan Chinese & \textit{gl} & Galician & \textit{gu} & Gujarati & \textit{he} & Hebrew \\ 
        \textit{hi} & Hindi & \textit{hr} & Croatian & \textit{hu} &  Hungarian & \textit{hy} & Armenian \\ 
        \textit{ia} & Interlingua & \textit{id} & Indonesian & \textit{is} & Icelandic & \textit{it} & Italian \\ 
        \textit{ja} & Japanese & \textit{jv} & Javanese & \textit{ka} & Georgian & \textit{kk} & Kazakh\\ 
         \textit{kn} & Kannada & \textit{ko} & Korean & \textit{ku} & Kurdish & \textit{la} & Latin \\ 
        \textit{lb} & Luxembourgish & \textit{lt} & Lithuanian & \textit{lv} & Latvian & \textit{mk} & Macedonian\\ 
         \textit{ml} & Malayalam & \textit{mn} & Mongolian & \textit{mr} & Marathi & \textit{ms} & Malay \\ 
        \textit{my} & Burmese & \textit{nds} & Low German & \textit{ne} & Nepali & \textit{nl} & Dutch \\ 
        \textit{nn} & Norwegian Nynorsk & \textit{no} & Norwegian & \textit{oc} & Occitan & textit{pl} & Polish 
        \\textit{pt} & Portuguese & \textit{ro} & Romanian & \textit{ru} & Russian & \textit{scn} & Sicilian \\ 
        \textit{sco} & Scots & \textit{sh} & Serbo-Croatian & \textit{si} & Sinhala & \textit{sk} & Slovak \\ 
        \textit{sl} & Slovenian & \textit{sq} & Albanian & \textit{sr} & Serbian &\textit{sv} & Swedish \\ 
        \textit{sw} & Swahili & \textit{ta} & Tamil & \textit{te} & Telugu & \textit{th} & Thai \\ 
        \textit{tl} & Tagalog & \textit{tr} & Turkish & \textit{tt} & Tatar & \textit{uk} & Ukrainian \\ 
         \textit{ur} & Urdu & \textit{uz} & Uzbek & \textit{vi} & Vietnamese & \textit{war} & Waray \\ 
         \textit{wuu} & Wu Chinese  & \textit{yi} & Yiddish & \textit{zh} & Chinese \\
        \textit{zh\_classical} & Classical Chinese & \textit{zh\_yue} & Cantonese &  & \\ 
         \bottomrule
    \end{tabular}
    \caption{Languages supported by BELA.}
    \label{tab:languages_table}
\end{table*}

\begin{figure*}[ht!]
\centering
     \includegraphics[clip,width=\textwidth]{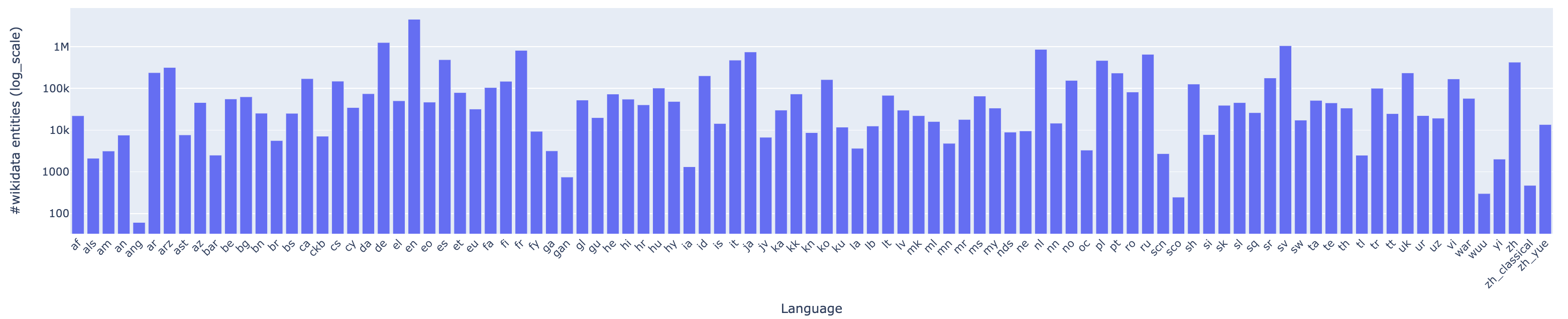}
\caption{Number of WikiData entities encoded in each language.}
\label{fig:entity_lang}
\end{figure*}

\section{Languages}
\label{sec:lang}
The 97 languages supported by BELA are listed in Table ~\ref{tab:languages_table}. They consist of the 99 languages supported by XLM-R, with the exception of \textit{zh\_min\_nan} and \textit{ceb}. Figure~\ref{fig:entity_lang} shows the histogram of languages chosen for encoding the WikiData entities.

\label{Hyperparameters}

\section{Data}
We show data statistics in Table \ref{tab:data_stats}. Note that we split Mewsli-9 in a training and a test (Mewsli-9') set. Our codebase lists respective indices for reproducibility.
\begin{table}[t]
\centering
\small
\begin{tabular}{lcc}
\toprule
\textbf{Dataset} & \textbf{\# Samples} & \textbf{Mentions} \\
\midrule
Mewsli-9& 58,717 & 289,087 \\
Mewsli-9'-train & 22,005 & 114543 \\
Mewsli-9'-validation & 7,351 & 32,383 \\
Mewsli-9'-test & 29,361 & 142,161 \\
Mewsli-9'-train (exhaustively labeled) & 22,005 & 437,862 \\
Mewsli-9'-validation (exhaustively labeled) & 7,351 & 134,241 \\
Mewsli-9'-test (exhaustively labeled) & 29,361 & 554,076 \\
TAC-KBP2015 & 333 & 12,522 \\
LORELEI & 2243 & 13,954 \\
AIDA  & 230 & 4,485 \\
Wikipedia ED test set & 100000 & 637797 \\
\bottomrule
\end{tabular}
\caption{Data statistics.}
\label{tab:data_stats}
\end{table}

\section{Underlabelling of TACKBP2015}
\label{Underlabelling}
One reason that our precision is low on the TACKBP2015 dataset is due to underlabelling of the dataset, i.e. some mention not being annotated as entities, but that our model detects.
Here are some examples of entities detected by our model but that are not considered true positives because of underlabelling:

For instance \textit{The \textbf{Investigatory Powers Tribunal} today released a judgment declaring that some aspects [...]} is correctly detected as "Investigatory Powers Tribunal" (wikidata: Q6060729).
Or \textit{From \textbf{New York City} in the 70s, late 80s and early 90s[...]} is correctly tagged as "New York City" (wikidata Q60).

\section{Training complexity}
\label{train_complex}
All numbers are reported using 40GB A100 GPUs.

Recall that we start by training the entity encoder in two steps.
In the first step (in-batch random negatives), we use 64 GPUs and train for approximately 4 days with a batch size of 5,120 (around 115k steps). In the second step (hard negatives),
we use 16 GPUs and train for 2 days with a batch size 960 (around 60k steps).

Next, to train the end-to-end model we again follow a two-step process. The disambiguation task trained on Wikipedia takes 5 days using 32 GPUs with a batch size of 24. The final end-to-end training takes 2 days using 32 GPUs with a batch size of 768.

Training the model from scratch is costly, but BELA's fast inference time makes it a suitable tool to integrate in large scale NLU systems.

\end{document}